\documentclass[journal]{IEEEtran}
\usepackage{cite}
\usepackage{graphicx}
\ifCLASSINFOpdf
\else
\fi
\usepackage{amsmath}
\usepackage{amsfonts}
\usepackage{amssymb}
\usepackage{algorithmic}
  \usepackage[caption=false,font=normalsize,labelfont=sf,textfont=sf]{subfig}
\usepackage{array}
\newcolumntype{x}[1]{>{\centering\arraybackslash\hspace{0pt}}p{#1}}

\begin{document}
% \title{Learning Flexible and Reusable Traffic Primitives of Driving Encounters}v1
% \title{Learning Traffic Primitives of Driving Encounters} %v2
% \title{Why Driving Can be Programmed? Learning the Building Blocks of Encounters}%v3
% \title{On Learning Fundamental Components When Encounters Other Vehicles}%v4
%\title{Learning the Building Blocks of Driving in Vehicle-Vehicle Encounters }%v5
%\title{Learning the Traffic Primitive Patterns of Driving in Vehicle-Vehicle Encounters }

\title{Understanding V2V Driving Scenarios through Traffic Primitives}
% \title{Learning the Primitives of Vehicle-Vehicle Encounters - The LEGO Bricks of Driving}%v6


\author{Wenshuo~Wang,~%\IEEEmembership{Student Member,~IEEE,}
        Weiyang~Zhang,~%\IEEEmembership{Fellow,~OSA,}
        and~Ding~Zhao%,~\IEEEmembership{Life~Fellow,~IEEE}% <-this % stops a space
\thanks{This manuscript was received by xxx.}
\thanks{Wenshuo Wang is with the Department of Mechanical Engineering, Carnegie Mellon University (CMU), Pittsburgh, PA 15213 USA. He was with the Department
of Mechanical Engineering, University of Michigan, Ann Arbor, MI 48109, USA. e-mail: wwsbit@gmail.com.}% <-this % stops a space
\thanks{Weiyang Zhang is with the Department of Mechanical Engineering, University of Michigan, Ann Arbor, MI 48109, USA. e-mail: zhangwy@umich.edu.}% <-this % stops a space
\thanks{Ding Zhao is with the Department of Mechanical Engineering, Carnegie Mellon University (CMU), Pittsburgh, PA 15213 USA. He was also with the Department of Mechanical Engineering, University of Michigan, Ann Arbor, MI 48109, USA. e-mail: dingzhao@cmu.edu.}
}

\maketitle

\begin{abstract}
Semantically understanding complex drivers' encountering behavior, wherein two or multiple vehicles are spatially close to each other, does potentially benefit autonomous car's decision-making design. This paper presents a framework of analyzing various encountering behaviors through decomposing driving encounter data into small building blocks, called driving primitives, using nonparametric Bayesian learning (NPBL) approaches, which offers a flexible way to gain an insight into the complex driving encounters without any prerequisite knowledge. The effectiveness of our proposed primitive-based framework is validated based on 976 naturalistic driving encounters, from which more than 4000 driving primitives are learned using NPBL -- a sticky HDP-HMM, combined a hidden Markov model (HMM) with a hierarchical Dirichlet process (HDP). After that, a dynamic time warping method integrated with $k$-means clustering is then developed to cluster all these extracted driving primitives into groups. Experimental results find that there exist 20 kinds of driving primitives capable of representing the basic components of driving encounters in our database. This primitive-based analysis methodology potentially reveals underlying information of vehicle-vehicle encounters for self-driving applications.
\end{abstract}

\begin{IEEEkeywords}
Driving scenarios, vehicle-to-vehicle, nonparametric Bayesian learning, traffic primitives.
\end{IEEEkeywords}

% \fi
\IEEEpeerreviewmaketitle

\section{Introduction}
\IEEEPARstart{D}{riving} encounter in this paper is referred to as the scenario where two or multiple vehicles are spatially close to and interact with each other\cite{wang2018clustering}, which occurs frequently in real traffic, for example, crossing at traffic intersections, merging into on-ramp highways, and changing lanes, etc., as shown in Table \ref{table:references}. The complete process of typical driving encounters usually consists of perception, decision-making, and control. In order to ensure traffic safety and efficiency, human drivers should precisely percept and be aware of their current surroundings, and seamlessly interact with nearby traffic participants. In driving encounters, human driving task in nature is a coupled dynamic and stochastic process\cite{nechyba1998stochastic}. For autonomous vehicles, making decisions when encountering with human drivers is challenging due to uncertainties on the continuous state of nearby vehicles and their potential discrete states such as changing lanes or turning at intersections. Likewise, in order to make autonomous vehicles capable of friendly and efficiently driving among human drivers, the interaction and decision-making processes of human driver, usually described by combination of continuous and potential discrete states, should be carefully investigated. 

\begin{table*}[t]
	\centering
	\caption{Existing State-of-the-art Research of Driving Encounter Modeling and Analysis via Vehicle Trajectories}
    \label{table:references}
	\begin{tabular}{cccccc}
		\hline\hline
		Year & Reference & Structured & Processing method & Scenario & Tasks\\
		\hline
		 2017 & \cite{sarkar2017trajectory} & $\checkmark$ & Hand-tuned heuristics & T-shape signalized intersections  &Crossing intersection \\
         2017 & \cite{galceran2017multipolicy} & & Bayesian change point & Cross-intersections; lane change & Multipolicy decision-making\\
         2016 & \cite{tang2016modeling} & $\checkmark$ &HMM & T-shape intersections& Decision-making\\
         2015& \cite{gindele2015learning} & $\checkmark$ &POMDP & Cross-intersection & Decision-making\\
         2012 & \cite{kasper2012object} & $\checkmark$ &Bayesian networks & Lane change & Behavior prediction\\
%         &&&highway on-ramps& Traffic merging\\
		\hline\hline
	\end{tabular}
\end{table*}

In order to formulate the relationship of continuous and potential discrete states, different approaches have been introduced and developed. One of the most flexible and popular ways is to use hybrid-state systems with decisions being model as a discrete-state system and dynamics of each state being model as a continuous-state. For example, some researchers utilized Gaussian mixture models (GMM) to account for nonlinearity and distribution of discrete decision states\cite{gadepally2014framework,havlak2014discrete,wang2018learning,butakov2015personalized} and then combined with a stochastic process model such as hidden Markov models (HMM); however, these approaches do not consider the history of the previous state when learning discrete states.
Dynamic Bayesian networks (DBN) also have been deployed for behavior prediction in lane change scenarios \cite{kasper2012object} and at cross-intersection\cite{gindele2015learning}, where the discrete states were empirically determined or learned from observations by maximizing specific expectation, which usually requires structured road conditions or contextual traffic. For example, Kasper, {\it et al.} \cite{kasper2012object} integrated lane-related coordinated systems with individual occupancy schedule grids based on DBN. Similarly, Gaussian mixture regression has been used to determine typical driving pattern for classification and prediction vehicle's trajectory on highways and at intersections \cite{lefevre2015driver,wang2018learning,tran2014online}. For all these aforementioned approaches, determining the potential discrete states is prerequisite, and most of them were deduced subjectively and empirically based on their prior knowledge such as traffic rules and lane boundaries\cite{zhu2015linear}. It is easy to apply in specified simple scenarios, but not for complicated scenarios with uncertainties of nearby vehicles such as those with agents not abiding by traffic rules.

Markov decision processes (MDP) provides a mathematically rigorous formalization of the decision-making problem for agents in driving encounters under uncertain conditions. For example, partially observable MDP (POMDP) \cite{gindele2015learning} was employed to model the internal relationship of the agents in driving encounters with a specifically predefined structured contextual traffic such as lane information (e.g., lane number and lane relations). Finding an optimal solution at a low computational cost for most POMDPs is still challenging, although some researchers have made remarkable contributions in methodologies and applications. For example, Ulbrich, et al. \cite{ulbrich2013probabilistic} applied an approximated way to find solutions of POMDP and Wei, et al.\cite{wei2011point} developed a point-based MDP by only taking uncertainty into account at the first planning step in single-lane merging scenarios. It will cause greatly expensive computations for POMDP when directly draw samples over full probability space such as action or/and state space. One way to ensure computational tractability is to reduce space size through drawing samples from a high-level policy space instead of from each data input space. Carefully segmenting action or state into subspaces could reduce the cost of drawing samples and hence ensuring computational tractability\cite{galceran2017multipolicy}.

According to aforementioned discussion and analysis, a well-built model of driving encounters should be able to adapt potential discrete states and describe the dynamics of each discrete state. However, diversity and complexity of driving contexts and human driver behavior make it challenging to select a mathematically rigorous approach capable of directly modeling and analyzing driver's interaction behavior, including both continuous and discrete states of all agents in driving encounters. Moreover, with the advances in data collection, the flood of high-dimensional and large-scale driving data can also overwhelm human mind\cite{wang2017much}.

%\subsection{Contributions}
In order to model drivers' dynamic interaction behavior (potential discrete and continuous states) in driving encounters based on vehicle trajectories without other contextual traffic and road information, this paper presents a driving primitive-based framework, which can gain insights into the basic components of driving encounters. The main contributions of this paper are three-fold: 1) introducing a driving primitive-based framework to investigate driving encounters using NPBL, 2) providing a clustering approach for primitives of driving encounters over the spatial and temporal spaces, and 3) demonstrating and analyzing the basic components of driving encounters based on naturalistic driving data.

The reminder of this paper is organized as follows. Section II introduces the driving primitive-based framework for analyzing driving encounters. Section III shows the experiment and data collection. Section IV discusses and analyzes the experimental results. Section V makes final conclusions of this work.

%\subsection{Paper Organization}

\section{Driving Primitive Extraction and Clustering}
\begin{figure}[t]
\centering
\includegraphics[width=\linewidth]{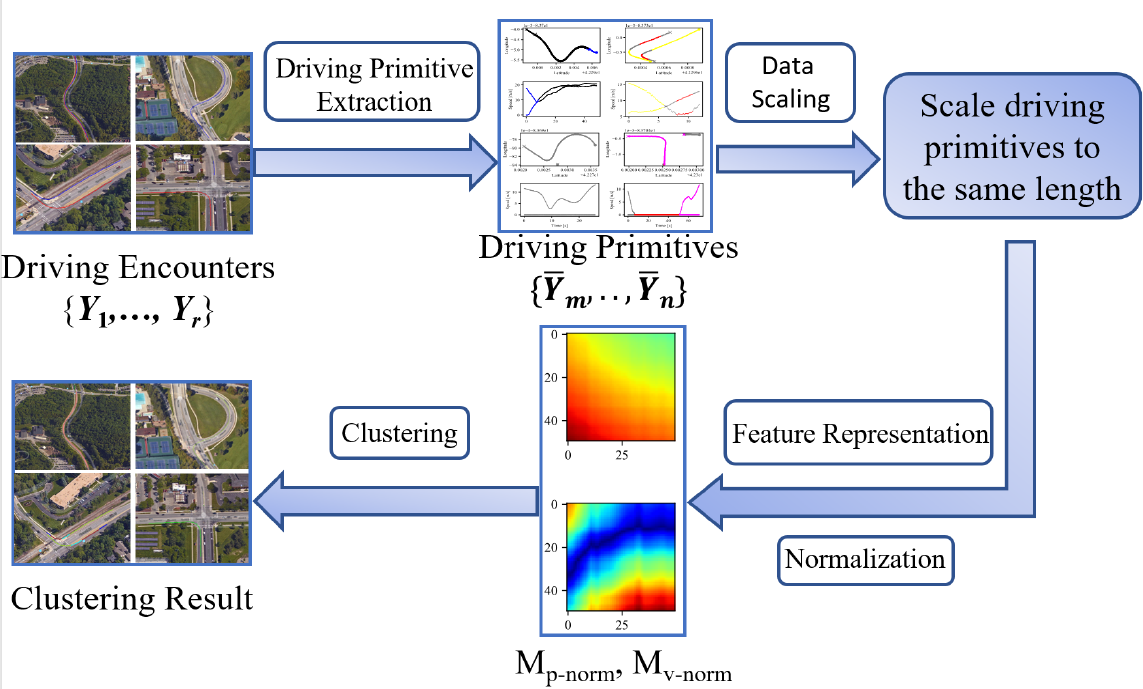}
\caption{Diagram of our proposed driving primitive-based framework.}
\label{fig:framework}
\end{figure}

In this section, we will introduce a methodology to extract primitives from driving encounters and then cluster these extracted primitives into groups. The problem will be mathematically formulated by three steps: primitive extraction, feature representation, and clustering, as shown in Fig. \ref{fig:framework}. 

\subsection{Driving Encounter}
In order to facilitate encountering behavior analysis and demonstrate the effectiveness of our proposed framework, we mainly focus on the driving encounter with two vehicle engaged. Our proposed methodology in Fig. \ref{fig:framework} could also be easily extended to the driving encounters with multiple vehicles engaged. The records of each vehicle consist of its position trajectories (latitude, longitude) and speeds, discretely ordered by time $t$. Thus, the driving encounter can be mathematically described as

\begin{equation}
\textbf{Y} = \{ y_1,\cdots, y_t, \cdots, y_{T} \}
%\textbf{\textit{X}} = \{ (p_1^{(1)}, p_1^{(2)}, v_1^{(1)}, v_1^{(2)}, t_1),...\ , (p_k^{(1)}, p_k^{(2)}, v_k^{(1)}, v_k^{(2)}, t_k) \}
\end{equation}
where $y_t = [p_t^{(1)}, p_t^{(2)}, v_t^{(1)}, v_t^{(2)}]\in \mathbb{R}^{6}$, $ p_t^{(1)} \in \mathbb{R}^2 $ and $ p_t^{(2)} \in \mathbb{R}^2 $ represent the position (latitude and longitude) of the first and second vehicles at time $t$, respectively; $v_{t}^{(1)}$ and $v^{(2)}_{t}$ are the speed of the first and second vehicles at time $t$, and $T$ is the length of data samples of the driving encounter $\textbf{Y}$. 

\subsection{Driving Primitive}
Driving primitives, in this paper, can be treated as the representation of fundamental building blocks of driving encounters. Each driving encounter can be segmented into finite numbers of primitives with distinct attributes. Each driving primitive does not have temporal overlap with others. The mathematical formulation of driving primitive is

\begin{equation}
\overline{\textbf{Y}} = \{ y_m,\cdots, y_n \}
%\overline{\textbf{\textit{X}}} = \{ (p_m^{(1)}, p_m^{(2)}, v_m^{(1)}, v_m^{(2)}, t_m),...\ , (p_n^{(1)}, p_n^{(2)}, v_n^{(1)}, v_n^{(2)}, t_n) \}
\end{equation}
where $\overline{\textbf{Y}} \subseteq \textbf{Y}$ with $m \leq n\leq T$. All the data falling within [$y_{m}$, $y_{n}$] is treated as the samples in this primitive. Theoretically, the length of driving primitives of a single driving encounter should be ($n-m+1$) $\in$ [1, $T$]. Since the driving primitive with very short duration provides little meaningful information, here we basically consider the driving primitive with long duration. 

\subsection{Driving Primitive Extraction}
In real traffic, analyzing driving encounters with high dimensional data is very difficult and beyond human mind. Usually we do not have rich enough prior knowledge of the types and amounts of driving primitive patterns. Although some mature classification methods such as support vector machines \cite{ben2001support,wang2017drivingstyle} and distance-based measures are able to capture the spatial attributes of time-series data, but failed to consider the temporal dynamic processes. To enable driving encounter analysis with high dimensional data tractable, we implement a nonparametric Bayesian learning approach -- the sticky HDP-HMM by combining a hierarchical Dirichlet process (HDP) with a hidden Markov model (HMM)\cite{fox2011sticky}, which can automatically segment time series into small pieces without requiring prior knowledge. In what follows, we will introduce the basics of the sticky HDP-HMM, which is derived based on HMM.

\subsubsection{HMM}
The general HMM usually consists of two layers: hidden layer and observed layer. Given an individual driving encounter with time-series data $\textbf{Y} = \{y_{t}\}_{t=1}^{T} \in \mathbb{R}^{6\times T}$ and a set of hidden state with $\mathcal{X}$, each hidden state at time $t$ will be subject to an entry of $\mathcal{X}$, i.e., $x_{t} = x_{i} \in \mathcal{X}$, where $x_{i}$ is the $i$-entry in $\mathcal{X}$. The transition probability from states $x_{i}$ to $x_{j}$ is denoted as $\pi_{i,j}$ with $\pi_{i} = [\pi_{i,1}, \pi_{i,2}, \cdots]$. The observation $y_{t}$ at time $t$ given hidden state $x_{t}$ is generated by $y_{t} = F(y_{t}|x_{t}, \theta_{x_t})$, called emission function, where $\theta_{t}$ is the emission parameter. Thus, the HMM model in our case can be described by

\begin{equation}
\begin{split}
x_{t}|x_{t-1}&\sim \pi_{x_{t-1}}\\
y_{t}|x_{t} &\sim F(\theta_{x_{t}})
\end{split}
\end{equation}
Then all the adjacent data samples in the driving encounter with the identical entry in $\mathcal{X}$ create a driving primitive $\overline{\textbf{Y}}$.

\subsubsection{HDP} Usually the number of driving primitives highly depends on the duration or the number of data samples of the driving encounter, that is, driving encounters with a longer duration will generate more driving primitives. In order to ensure the transition probability $\pi_{i,j}$ always keeps $\sum \pi_{i,j} = 1$ when feeding new data samples into the model, we consider the time-series data of encountering behavior as a Dirichlet process (DP)\cite{teh2005sharing}, denoted by DP$(\gamma, H)$, which can be formulated by

\begin{subequations}
\begin{equation}
G_0 = \sum_{i=1}^{\infty}\beta_i\delta_{\theta_i},\       \ \theta \sim H
\end{equation}
\begin{equation}
\beta_i=v_i\prod_{\ell=1}^{i-1}(1-v_\ell),\       \ v_i \sim Beta(1,\gamma)
\end{equation}
\end{subequations}
where $\gamma$ is the hyperparameter and $\beta_i$ are the weights  sampled by a stick-breaking construction, denoted as $\beta\sim$ GEM$(\gamma)$, which ensures the sum of weights $\beta_{i}$ is always equal to one even for infinite amounts of data samples. The prior of the HMM transition probability measures $G_{j}$ is defined by a hierarchical DP (HDP) and formulated as

\begin{equation}
G_{j}=\sum_{i=1}^{I}\pi_{j,i}\delta_{\theta_i}
\end{equation}
where $\delta_\theta$ is a mass concentrated at $\theta$. In our case, we take $G_j \sim DP(\alpha, G_0)$, where $G_0$ is drawn by DP.

\begin{figure}[t]
\centering
\includegraphics[width=0.5\linewidth]{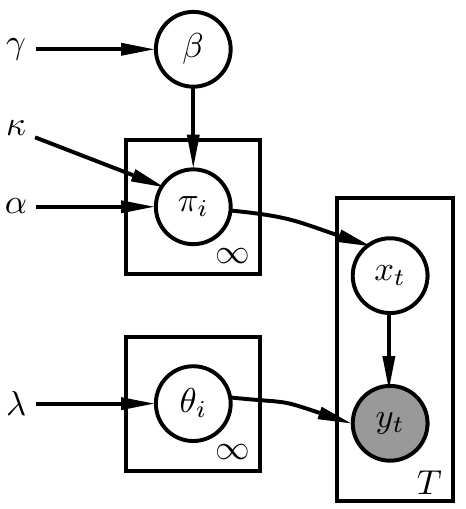}
\caption{Illustration of the sticky HDP-HMM approach.}
\label{fig:sticky_hdp_hmm}
\end{figure}

Based on aforementioned model, we finally introduce the stick HDP-HMM \cite{fox2011sticky,wang2017driving,wang2018extracting} to control the expected self-transition probability by adding a parameter $k\in [0,1]$. The graphic illustration is shown in Fig. \ref{fig:sticky_hdp_hmm}. Therefore, we can conclude the whole method as

\begin{equation}
\begin{split}
\beta|\gamma &\sim GEM(\gamma) \\
\pi_i|\alpha,\beta,k &\sim DP(\alpha+k, \frac{\alpha\beta+k\delta_i}{\alpha+k}) \\
x_t|x_{t-1} &\sim \pi_{x_{t-1}}\\
y_t|x_t & \sim F(\theta_{x_t})\\
\theta_i|H &\sim H
\end{split}
\end{equation}
% \begin{subequations}
% \begin{equation}
% \beta|\gamma \sim GEM(\gamma) \\
% \end{equation}
% \begin{equation}
% \pi_i|\alpha,\beta,k \sim DP(\alpha+k, \frac{\alpha\beta+k\delta_i}{\alpha+k})
% \end{equation}
% \begin{equation}
% x_t|x_{t-1} \sim \pi_{\overline{\textbf{\textit{X}}}_{t-1}}
% \end{equation}
% \begin{equation}
% y_t|x_t \sim F(\theta_{x_t})
% \end{equation}
% \begin{equation}
% \theta_i|H \sim H
% \end{equation}
% \end{subequations}
where $t\in[0, T]$. All the hyperparameters are set as a Gamma distribution for the convenience of estimating the posterior probability of hidden states. For the emission parameter $\theta$, in our case, we assume the emission model is subject to a Gaussian distribution as in \cite{wang2017driving}, which makes the posterior estimation more tractable with Gibbs sampling algorithms.

\subsection{Feature Representation}
%Feature representation is an essential step for data processing. 
In order to make further reuse and analysis easily, we need classify these extracted driving primitives. Feature representation is crucial for classifying time-series data. We shall extract expected features able to capture the similarity between two driving primitives in both temporal and spatial spaces without losing much information of the raw time-series data. Measuring the distance between two time-series data is the general idea of similarity evaluation. Some methods have been developed such as the Euclidean distance with mean values, Pearson's correlation coefficient with cross-correlation-based distances, short time series distance (STS)\cite{moller2003fuzzy}, dynamic time warping\cite{muller2007dynamic}, and probability-based distance function. More details of these methods can refer to survey paper \cite{liao2005clustering}. Since rare prior knowledge on the feature space of driving primitives is available, we prefer to consider the  spatial and temporal distances of any two driving primitives, which prompts us to implement the dynamic time warping (DTW) method.
% * <wwsbit@gmail.com> 2018-06-10T13:10:40.576Z:
% 
% > Since rare prior knowledge of the feature space of driving primitive is available, we prefer to keep enough feature used for clustering, which is suitable to implement dynamic time warping.
% Not so convincible.
% 
% ^.

DTW is one common way to measure the geometrical similarity of two time-series data with the same length. The position trajectories and speeds of the two engaged vehicles for individual driving encounters are employed equivalently to describe driving primitives. The DTW method allows us to get the corresponding spatial and temporal feature matrices by using position trajectories and speeds, respectively, which is easily operated. Given one driving primitive $\overline{\textbf{Y}} = \{ p^{(1)}, p^{(2)}, v^{(1)}, v^{(2)} \}$ with a length of $n-m+1$, where

\begin{subequations}
\begin{equation}
p^{(1)} = [ p_m^{(1)}, p_{m+1}^{(1)},...\ , p_n^{(1)} ]
\end{equation}
\begin{equation}
p^{(2)} = [ p_m^{(2}, p_{m+1}^{(2)},...\ , p_n^{(2)} ]
\end{equation}
\begin{equation}
v^{(1)} = [ v_m^{(1}, v_{m+1}^{(1)},...\ , v_n^{(1)} ]
\end{equation}
\begin{equation}
 v^{(2)} = [ v_m^{(2}, v_{m+1}^{(2)},...\ , v_n^{(2)} ]
\end{equation}
\end{subequations}
we denote the feature matrices of position trajectory and speed are denoted as $M_p$, $M_v \in \mathbb{R}^{(n-m+1)\times(n-m+1)}$, respectively. Considering $P_{i,j}$ as the elements of matrices $M_p$ and $V_{i,j}$ as the elements of matrices $M_v$ with $P_{i,j}, V_{i,j}\in \mathbb{R}_{\geq0}$ and $m\leq i,j \leq n$, then the element $P_{i,j}$ represents as the local measure between $p_i^{(1)}$ and $p_j^{(2)}$, and the element $V_{i,j}$ represents as the local measure between $v_i^{(1)}$ and $v_j^{(2)}$. In order to evaluate local measures $P_{i,j}$ and $V_{i,j}$ quantitatively, we reformulate them by

\begin{subequations}
\begin{equation}
p_{i,j} =: D_p(p_i^{(1)}, p_j^{(2)})
\end{equation}
\begin{equation}
v_{i,j} =: D_v(v_i^{(1)}, v_j^{(2)})
\end{equation}
\end{subequations}
where $D(\cdotp,\cdotp)$ is a distance measure function. Since $p \in \mathbb{R}^2$ is a two dimensional data point and $v \in \mathbb{R}$, then we choose the Euclidean distance to calculate $D_p$

\begin{subequations}
\begin{equation}\label{eq:Dp}
D_p(p_i^{(1)}, p_j^{(2)}) = \|p_i^{(1)}, p_j^{(2)}\|_2
\end{equation}
and the Manhattan distance to calculate $D_v$
\begin{equation}\label{eq:Dv}
D_v(v_i^{(1)}, v_j^{(2)}) = \|v_i^{(1)}, v_j^{(2)}\|_1
\end{equation}
\end{subequations}
By using DTW, we can obtain a interpretable feature representation, for example, a smaller $D_p$ value indicates the positions of the two vehicles are spatially close to each other and a smaller $D_v$ value indicates smaller speed difference between the two vehicles.  %By using DTW, we can both save the geometrical features and discrete time sequence by the elements' local distance and location in the matrices.

\subsection{Scaling and Normalization}
Since the durations of the driving primitives extracted by the sticky HDP-HMM are greatly different, directly forming the feature representation without any size-scaling using (9) could make clustering tasks intractable. Therefore, scaling the driving primitives into the identical length is a prerequisite step for getting applicable features. In addition, the magnitude of position trajectory data recorded from GPS is totally different from that of speed, which would generate unbalanced feature representations, thus leading to unexpected clustering results. To overcome these problems, we need to up-scale and down-scale the length of data and normalize the feature matrices.

\subsubsection{Length scaling}
We implement the linear interpolation\cite{meijering2002chronology} to scale the time series data into a predefined length. Considering two driving primitive vectors $y_0 =[ p_0, v_0]$ at time $t_0 $ and $y_1 =[p_1, v_1]$ at time $t_1$, the unknown driving primitive $x$ at time $t\in [t_0, t_1]$ can be approximated by

\begin{equation}
y_{t} = \left[p_0+(t-t_0)\frac{p_1-p_0}{t_1-t_0},\ v_0+(t-t_0)\frac{v_1-v_0}{t_1-t_0}\right]
\end{equation}

By applying (10), each driving primitive can be rescaled to the predefined length $l$, in which the derivative might be discontinuous.

\subsubsection{Normalization of Feature Matrices}
There exist lots of mature approaches to normalize data such as mean normalization, rescaling, standardization, division by maximum, etc. Since the units of trajectory have the different scale with speed, we should normalize $M_p$ and $M_v$ separately. In our case, each element in the feature matrices is nonnegative, therefore the method we can use is division by the maximum of feature matrices. In such way, we can avoid the situation when the minimum value is the same as the maximum value, which is common in practical driving behavior such as the vehicle is still. The formal formulation of feature normalization is

\begin{subequations}
\begin{equation}
M_{p-norm} = \frac{M_p}{\max(M_p)}
\end{equation}
\begin{equation}
M_{v-norm} = \frac{M_v}{\max(M_v)}
\end{equation}
\end{subequations}
where $\max(M_p)$ and $\max(M_v)$ are the maximum values of feature matrices in terms of position trajectories and speeds, respectively.

\subsection{Clustering}
Clustering the driving primitives with time-series data is challenging without any prior knowledge and ground truth. The details of different clustering methods are in review paper.\cite{xu2015comprehensive} Clustering can be classified as different methods based on different clustering criteria, such as $k$-means \cite{jain2010data}, BIRCH \cite{zhang1996birch}, DBSCAN \cite{guha1998cure}, etc. In this paper, we select the $k$-means clustering method to our case.

Given one driving primitive $\overline{\textbf{Y}}$ with the normalized trajectory feature matrix $M_{p-norm}$ and speed feature matrix $M_{v-norm}$, we reformat them into a vector by

\begin{subequations}
\begin{equation}
\begin{split}
\textbf{\textit{f}} : (M_{p-norm}, M_{v-norm}) \rightarrow \phi \\ 
\mathbb{R}^{l\times l} \times \mathbb{R}^{l\times l} \rightarrow \mathbb{R}^{1\times 2l^2}
\end{split}
\end{equation}
\begin{equation}
\phi = [ p_{1,1},...\ , p_{l,l}, v_{1,1},...\ ,v_{l,l} ]
\end{equation}
\end{subequations}
where $\phi$ is a primitive feature vector which covers all of the trajectory and speed features information. Given a primitive feature vector set $\boldsymbol{\phi} = \{\phi_1, \phi_2,... \, \phi_N \}$ with $N$ being the number of driving primitives, $\boldsymbol{\phi}$ can be partitioned as $k$ ($\leq N$) clusters set $C = \{C_1,...\ ,C_k \}$ by implementing the $k$-means clustering method to these feature vectors.  
Consider $\mu_i$ as the mean of cluster $C_i$, we aim to minimize the distance within clusters with the objective formulated by

\begin{equation}
\min \sum_{i=1}^k \sum_{\phi_i \in C_i} \| \phi_i - \mu_i \|^2
\end{equation}
By minimizing the sum of within-cluster's squared errors, the primitives with similar features (both trajectory and speed) can be assigned into same clusters. Since the ground truth is unavailable, evaluating the result of unsupervised learning can be challenging. Here, we introduce the with-in distance $\lambda_w$ and between distance $\lambda_b$ to evaluate clustering performance, described as

\begin{subequations}
\begin{equation}
\lambda_w = \frac{\sum_{i=1}^k \sum_{\phi_i \in C_i} \| \phi_i - \mu_i \|^2}{N-k}
\end{equation}
\begin{equation}
\lambda_b = \frac{\sum_{i=1}^k {n_i\|\mu_i - \bar{\mu}\|}^2}{k-1}
\end{equation}
\end{subequations}
where $n_i$ is the number of driving primitives in the cluster $i$, $\bar{\mu}$ is the mean of $\boldsymbol{\phi}$. Usually, with the cluster number $k$ increasing, both $\lambda_w$ and $\lambda_b$ will decrease. The cluster number can be determined by weighing computational cost and the decreasing speed of $\lambda_w$ and $\lambda_b$.

\section{Experiment and Data Collection}
In this section, we will introduce the data collection and preprocessing before  extracting driving primitives and taking further analysis, including data collection and extraction, data unification and normalization, and hyperparameter settings.

\subsection{Data Collection and Extraction}
The driving encounter data was collected from the database generated by the University of Michigan Safety Pilot Model Development (SPMD) program and conducted by the University of Michigan Transportation Research Institute (UMTRI)\cite{wang2018extracting}. The database includes approximately 3,500 equipped vehicles and 6 million trips in total for more than 3 years. Dedicated short range communications (DSRC) technology was applied for the communication between two vehicles, which would be activated when two vehicles are close to each other within 100 meters. That means the driving encounter is defined as the mutual behaviors when two vehicles' Euclidean distance is less than 100 meters. The latitude and longitude data of each vehicle was collected by the on-board GPS. Speed information was collected by the by-wire speed sensor. Data refresh frequency is 10 Hz.

In order to get meaningful and extensive information from the driving encounter behaviors, we focused on the driving encounters with duration of longer than 10 s. After acquiring and searching the database, 976 qualified driving encounters were extracted. %Each driving encounter includes the latitude, longitude, and speed of two vehicles.

\subsection{Data Scaling and Normalization}
\begin{figure}[t]
\centering
\includegraphics[width=\linewidth]{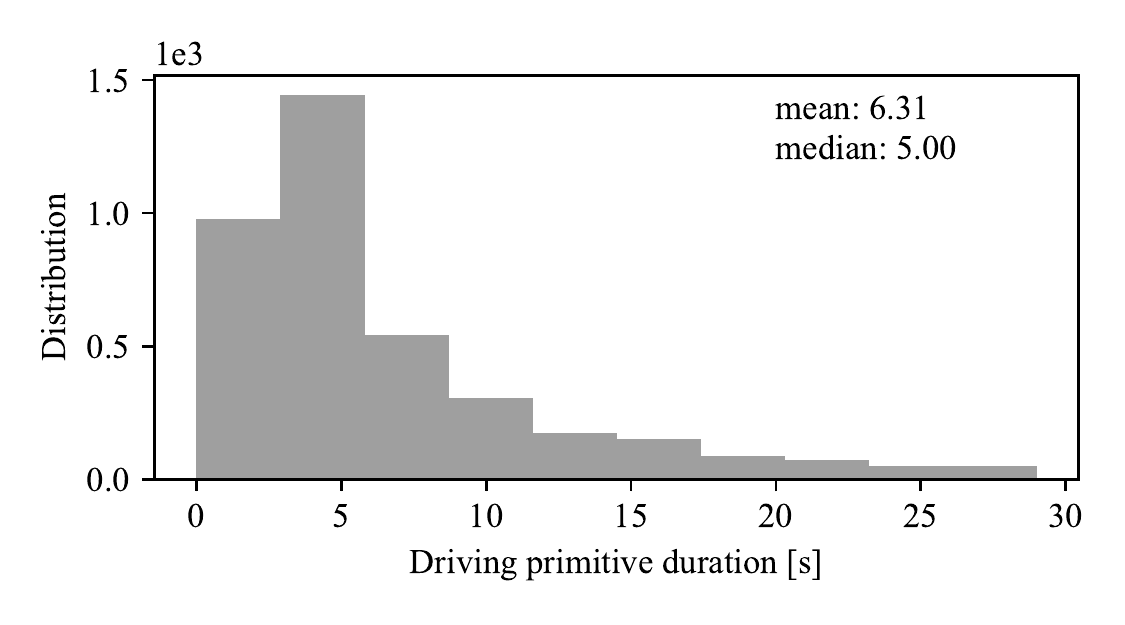}
\caption{Distribution of driving primitive duration. The mean duration of primitive is 6.31s. The median duration of primitive is 5.00s.}
\label{fig:durations}
\end{figure}

By applying the sticky HDP-HMM, 4126 driving primitives were generated from 976 driving encounters. Before taking clustering processes, we scaled the length of driving primitive and then normalized the value of the feature matrices, then obtained $M_{p-norm}$ and $M_{v-norm}$ for each driving primitive. Determining the scaling length for driving primitives is very important but hard. An overlarge length will enlarge the size of feature representation and then significantly increase the computational cost, while a too short length would loss some information of driving encounters. Fig. \ref{fig:durations} shows the distribution of the original driving primitives. It can been seen that most of the driving primitives have a length of around 5.00 s. In order to extract feature matrices with balancing the computational cost and losing less information, the expected primitive duration is set as 5.00 s in 10 Hz, i.e., with 50 data samples  ($l=50$) for each driving primitive.

%Considering the scale of changes in trajectory and speed, the change of latitude and longitude is much lower in magnitude than changes in speed. 
For data normalization, the local distance of each feature matrix (trajectory or speed) is calculated using (9) and then normalized using (11) which can keep features within the range of [$0$,  $1$]. After that, we then combined them together and reformatted into a feature vector through Equation (12).

\subsection{Experiment Parameter Settings}

%\textit{1) Sticky HDP-HMM model settings:} 
%We set the maximum type of driving primitives extracted from a single driving encounter as 10, which means one single driving encounter can extract at most 10 different kinds of driving primitives.
When training the sticky HDP-HMM, the iteration number for estimating posterior probability of latent states was set as 200. In order to implement linear interpolation to sample the extracted primitives into the predefined length, we only considered driving primitive whose duration is longer than 0.2 s. Driving primitive with very short time period can only provide limited useful information and may interfere with other meaningful information.    

\begin{figure*}[t]
\centering
\includegraphics[width=\linewidth]{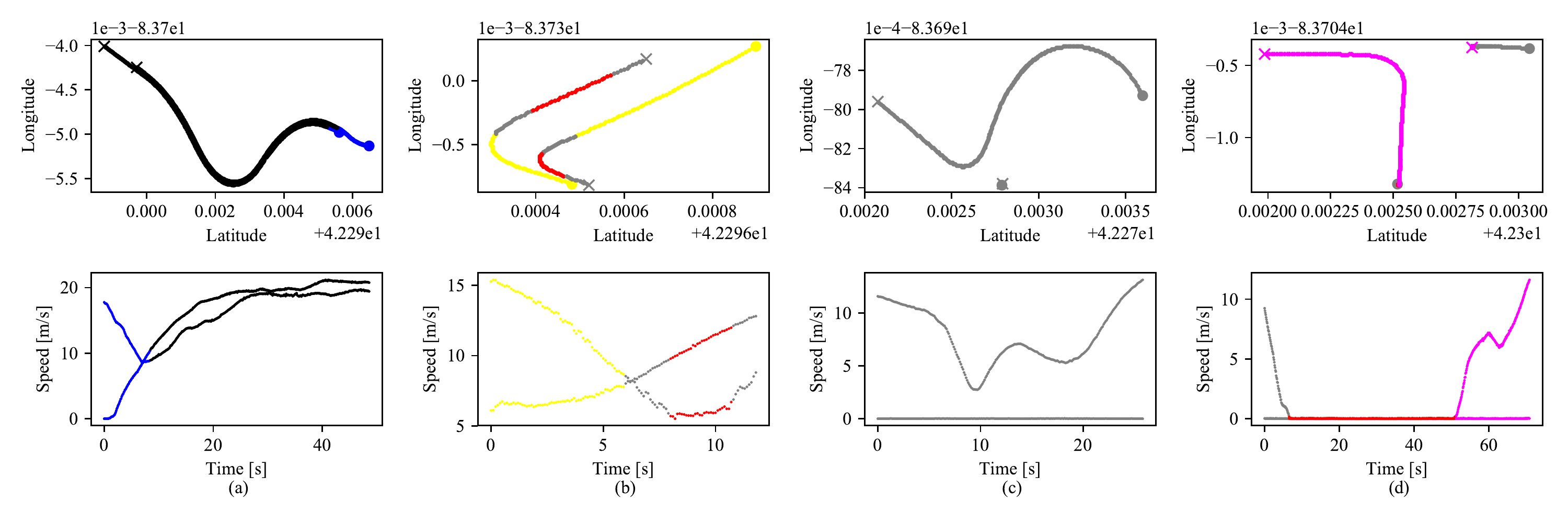}
\caption{Examples of four driving primitives extraction results, consisting of vehicle trajectories (top) and vehicle speed (bottom).}
\label{fig:original_encounter_result}
\end{figure*}

\begin{figure*}[t]
\centering
\includegraphics[width=\linewidth]{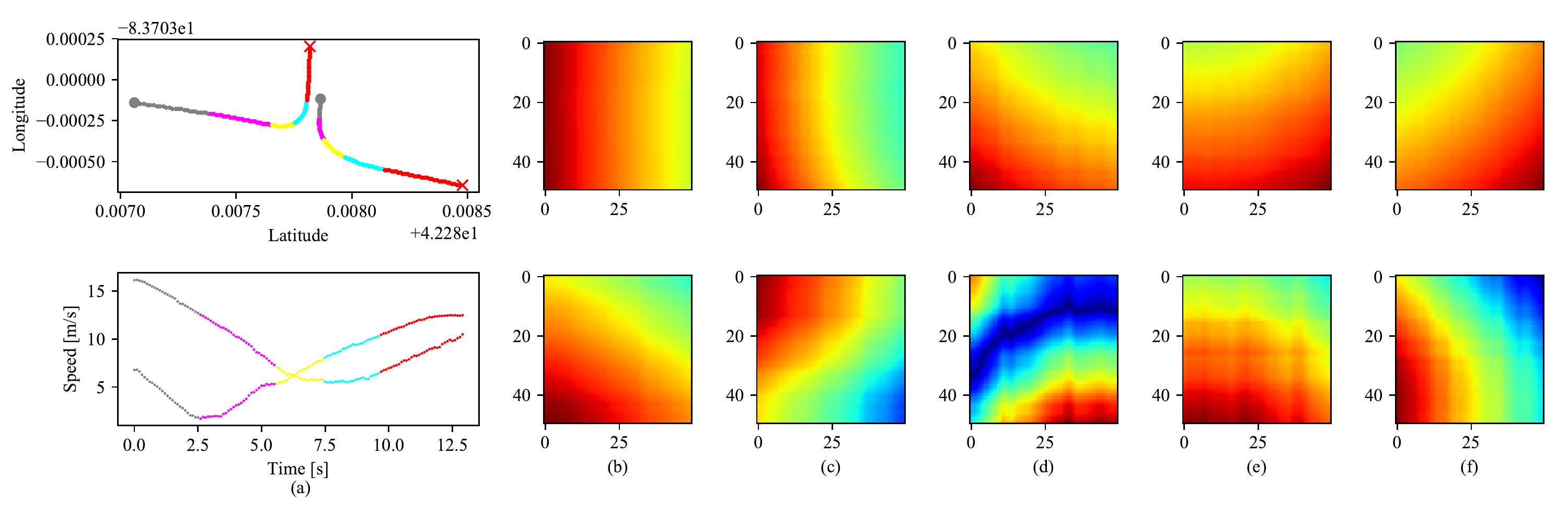}
\caption{Example of driving encounter with trajectories and speed feature matrices for different driving primitives. (a) shows the trajectory and speed of the whole driving encounter. (b), (c), (d), (e), (f) show feature matrices calculated by DTW of each driving primitive, which correspond to the gray, pink, yellow, blue and red colors in (a) respectively.}
\label{fig:DTW}
\end{figure*}

%\textit{2) DTW and rescaling:} 
Considering the length of driving primitive data, $M_p$ and $M_v$ feature matrices generated from DTW were both set as $50\times 50$ dimensions (i.e., $l=50$), which can be calculated by (\ref{eq:Dp}) and  (\ref{eq:Dv}). The dimension of primitive feature vector $\phi$ is $\mathbb{R}^{1\times 5000}$ after rescaling using (12).

\section{Results and Analysis}
This section will present the experiment results and analysis of driving primitive extraction and clustering.

%\begin{figure*}
%\centering
%\includegraphics[width=\linewidth]{figures/Figure_encounter_heatmap}
%\caption{The heat map of driving encounters. (a) All driving encounters. (b) Driving encounters after removing those which have same driving primitives.}
%\label{fig:encounter_heatmap}
%\end{figure*}

\subsection{Driving Primitive Extraction Result and Analysis}

Fig. \ref{fig:original_encounter_result} visualizes some examples of the extracted driving primitives of four typical driving encounters using the sticky HDP-HMM. The different colors represent different driving primitives. The dot and cross represent the starting point and endpoint of driving primitives, respectively. The results demonstrate that the sticky HDP-HMM can efficiently detect the boundaries of driving primitives for driving encounter time-series data and then label them. According to the experimental results, we can make key conclusions in three-fold as follows.

\begin{itemize}
\item Extracted driving primitives are in line with human subjective expectations, that is, the driving primitives extracted by the sticky HDP-HMM is relatively explainable. For instance, Fig. \ref{fig:original_encounter_result}(c) shows the typical encountering behavior that one vehicle moves and another vehicle stops while keeping engine on. In our case, the entire driving encounter behavior is considered as one primitive, which is compatible with human intuitive understanding as well as knowledge of driver behavior. Take Fig. \ref{fig:original_encounter_result}(d) for example, it represents a typical driving encounter behavior at the intersection, where one vehicle decelerates and then stops and waits for a red light, while another vehicle stops first and then turns left after the traffic signal turning green. The driving encounter is separated as three different primitives in our case: a) one vehicle decelerates and then stop  while another keeps still. b) Both two vehicles do not move. c) one vehicle does not move while another turns left.

\item Individual driving encounters may consist of different types and numbers of driving primitives. From Fig. \ref{fig:original_encounter_result}, we can see that each driving encounter includes more than one driving primitive, except for Fig. \ref{fig:original_encounter_result}(c). Especially for Fig. \ref{fig:original_encounter_result}(b), it shows two pieces of driving primitives are both in gray, which means they are considered as one type of driving primitive.

\item Speed information is a good complement to trajectory information. If merely trajectory information is implemented, there will exist feature information loss. For instance, in Fig. \ref{fig:original_encounter_result}(a), the trajectories of two vehicles are nearly overlapped, which represents the scene of two cars traveling on the road with the same direction. Without speed information, it can be challenging to separate the driving encounter as two primitives, such as it will be considered as one primitive like Fig. \ref{fig:original_encounter_result}(c), which indicates that speed information is necessary for primitive extraction.
\end{itemize}

In summary, the above experiment result and analysis demonstrate that the behavior of driving encounters can be decomposed into some basic interpretable components, called driving primitives. These extracted primitives then offer a flexible way to further analyze complex driving behavior with multiple agents engaged that beyond human understanding.

\begin{figure}[t]
\centering
\includegraphics[width=\linewidth]{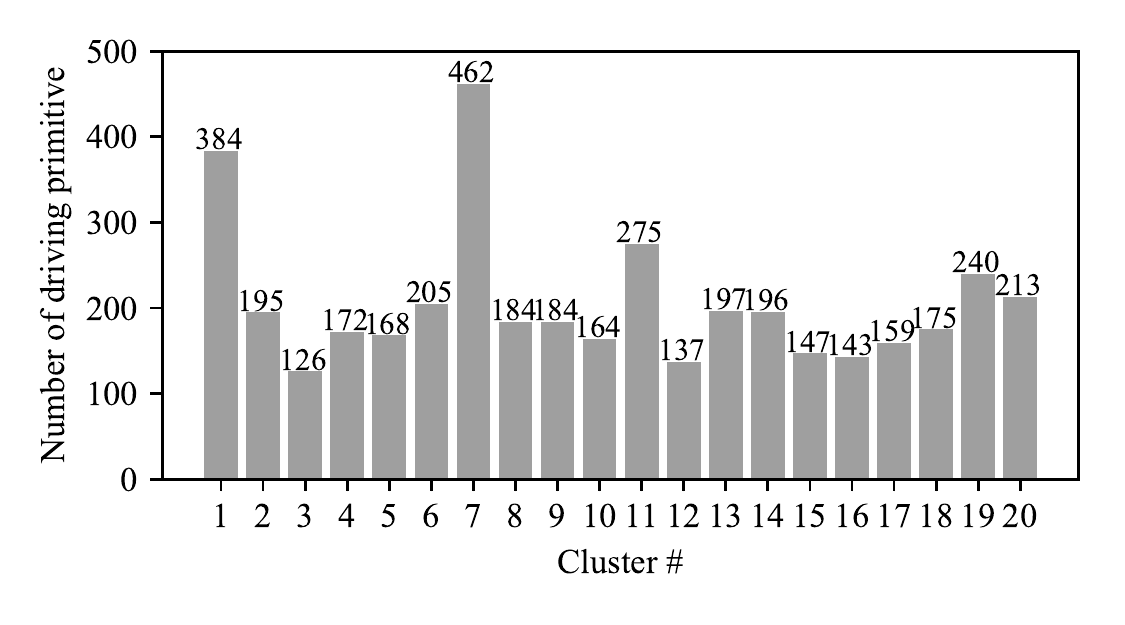}
\caption{Distribution of 20 clusters calculated using $k$-means clustering.}
\label{fig:primitive_num}
\end{figure}

\subsection{Driving Primitive Feature Analysis}
% * <zhaoding1014@gmail.com> 2018-06-10T16:04:40.779Z:
% 
% Weiyang and Wenshuo, I feel the conclusion part is a little bit week.... Could you add the Ann Arbor figure to the analysis part? Also, we may also want to include more discussions to the 20 primitives and their relationship to the encounters, statistically and by case-study.
% 
% ^.
In order to analyze and show the dynamic differences among driving primitives, 
we compute their DTW features. Fig. \ref{fig:DTW}(a) illustrates an example of common driving encounter occurred at the intersection, which was separated into five different driving primitives with different colors. Fig. \ref{fig:DTW}(b)--(f) are the color maps representing the feature matrices (top for position trajectories and bottom for speed) computed through DTW. The blue represents a short local distance and the crimson represents a long local distance. By visualizing these matrices using heat maps, the difference between driving primitives can be easily and intuitively captured. Take the speed feature matrix for example, the feature map (bottom) of Fig. \ref{fig:DTW}(d) is covered with large blue areas in the center and red at both ends of leading diagonal, which shows most distinct from the feature maps of other driving primitives in this driving encounter. That means in the middle of this time period area, the speed difference between the two vehicles is very small. Correspondingly, the speed difference between the two vehicles is large at both ends of this time period. The yellow part in bottom figure of Fig. \ref{fig:DTW}(a) displays the speed profiles of the two vehicles, which can support the conclusion mentioned above. During this driving primitive, the speed difference between two vehicles changes from large to small and then become large again. Besides, there exists a crossover of speed during this time period, which indicates a zero Manhattan distance, displayed as deep blue in the bottom plot of Fig. \ref{fig:DTW}(d).

Thus, it can be verified that the heat maps of feature matrices of both position trajectory and speed in Fig. \ref{fig:DTW} are different among individual driving primitives, which indicates that the spatial and temporal features of driving primitives can be reserved by implementing DTW. 

\subsection{Clustering Result and Analysis}
After applying the sticky HDP-HMM to driving encounter data, we obtained large amounts of driving primitives; however, it is still unknown about exactly how many kinds of driving primitives  compose these driving encounters. Here we are going to investigate the type of driving primitives in the view of the spatial and temporal features which can be captured through DTW. Based on the normalized feature matrices of position trajectory and speed, as shown in Fig. \ref{fig:DTW}, we then implemented $k$-means to cluster these feature matrices into groups, thus the correlated driving primitives can be grouped. $k$-means clustering is an unsupervised approach, therefore we need first to determine the number of clusters.

\begin{figure}[t]
\centering
\includegraphics[width=\linewidth]{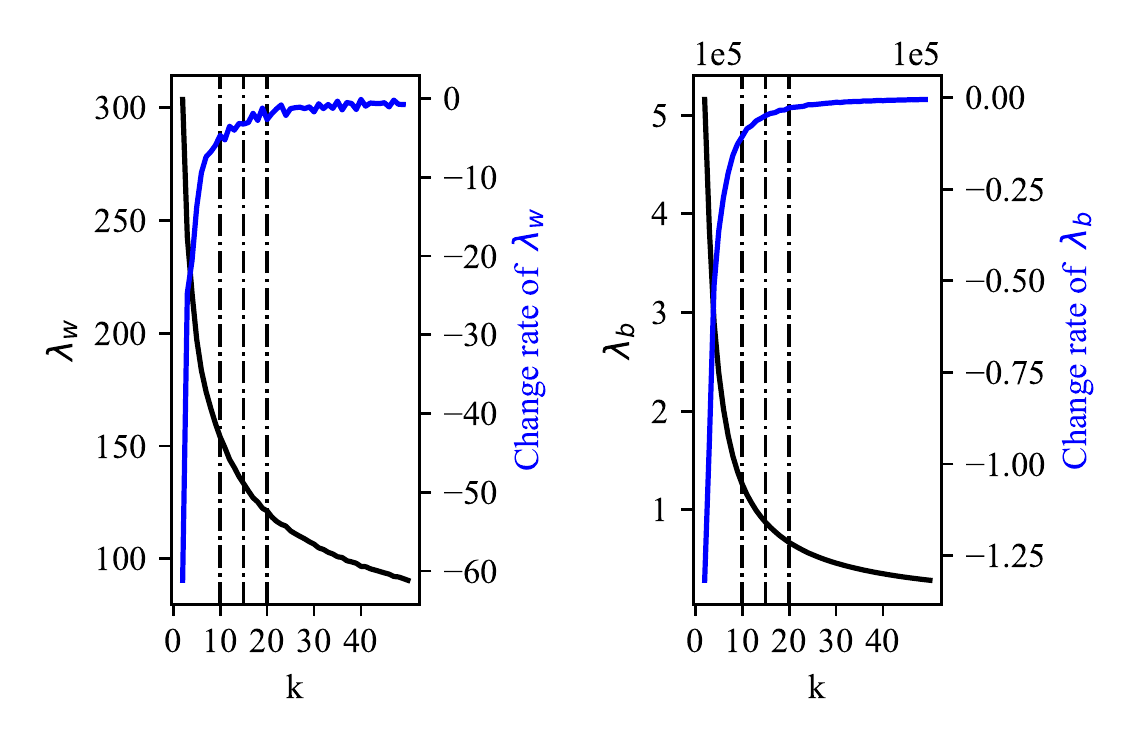}
\caption{Results of within distance $\lambda_w$ and between distance $\lambda_b$ over the number of clusters $k$.}
\label{fig:cluster_distance}
\end{figure}

Fig. \ref{fig:cluster_distance} shows the values of within distance $\lambda_w$ and between distance $\lambda_b$ and their change rates over the number of clusters $k$ varying from 2 to 50. It can be seen that both $\lambda_w$ and $\lambda_b$ decrease when increasing $k$, and their change rates trend to zero correspondingly. When the number of clusters $k$ reaches 20 ($k=20$), the change rates (blue line) of both $\lambda_w$ and $\lambda_b$ tend to converge. With considering the computational cost and model accuracy, we finally select $k=20$ at the `elbow' point of the change rate of $\lambda_w$ and $\lambda_b$ because selecting a very large number of clusters will suffer an excessive computational cost but without any significant performance improvement.

%\begin{figure*}
%\centering
%\includegraphics[width=\linewidth]{figures/Figure_gps_typical1-min1}
%\caption{GPS data visualized on Google map and four typically driving primitives with trajectories and speed feature matrices. (a) demonstrates all the driving primitives' trajectories with their clustering colors as well as a zoomed in intersection situation. (b) shows two most common (upper two plots) and uncommon (lower two plots) driving primitives with their feature matrices.}
%\label{fig:gps_plot}
%\end{figure*}

Fig. \ref{fig:primitive_num} displays the distribution of driving primitives in $k=20$ clusters. Each cluster contains more than $100$ driving primitives. The most common driving primitive in our case is the cluster \#$7$, accounting for 11.20\% of total driving primitives and the most rare appearance one is the cluster \#$3$, accounting for 3.05\% of total driving primitives. Each driving primitive represents the basic building blocks of driving encounters. The detailed clustering results enable us to further analyze the distribution of driving primitives in individual driving encounters. 

\begin{figure}[t]
\centering
\includegraphics[width=\linewidth]{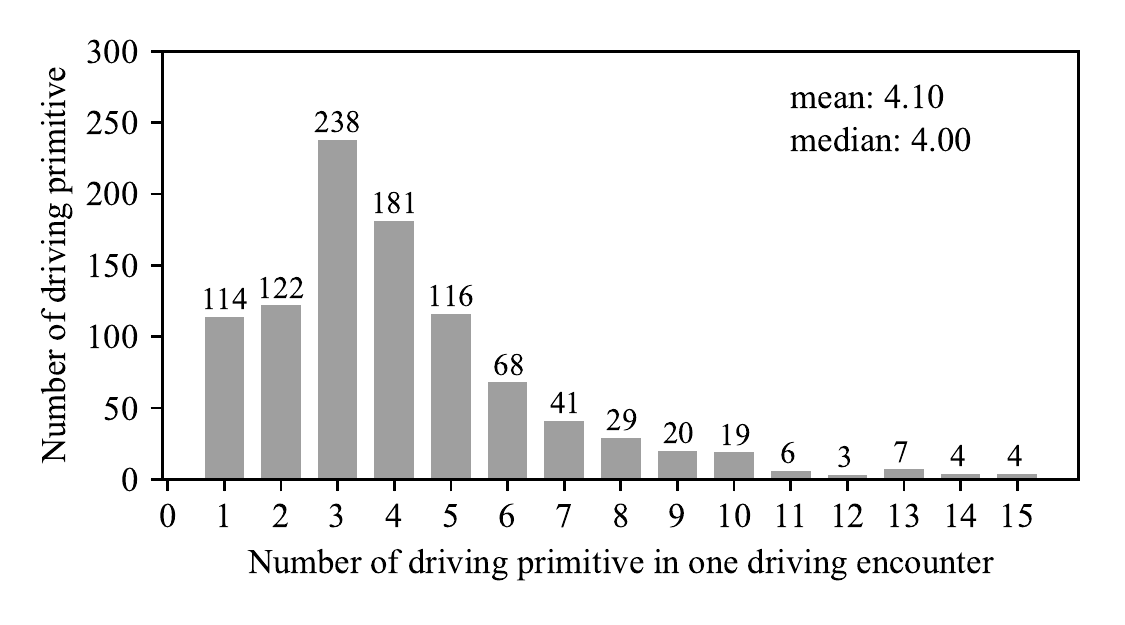}
\caption{Distribution of the number of driving primitives in individual driving encounters.}
\label{fig:primitive_num_encounter}
\end{figure}

Different driving encounters could be composed of different kinds and numbers of driving primitives. Fig. \ref{fig:primitive_num_encounter} displays the distribution of the number of driving primitives in individual driving encounters. It can be known that most driving encounters are composed of 3 driving primitives, accounting for 24.39\% of total driving encounters and few driving encounters are composed of more than 10 driving primitives, accounting for 2.5\% of total driving encounters. The mean and median of the number of driving primitives in individual driving encounters are both around 4. In addition, there exist 114 driving encounters which consist of only one driving primitive, that is, the driving encounter was considered as a driving primitive such as the driving encounter in Fig. \ref{fig:original_encounter_result}(c).

\begin{figure*}
\centering
\includegraphics[width=0.98\linewidth]{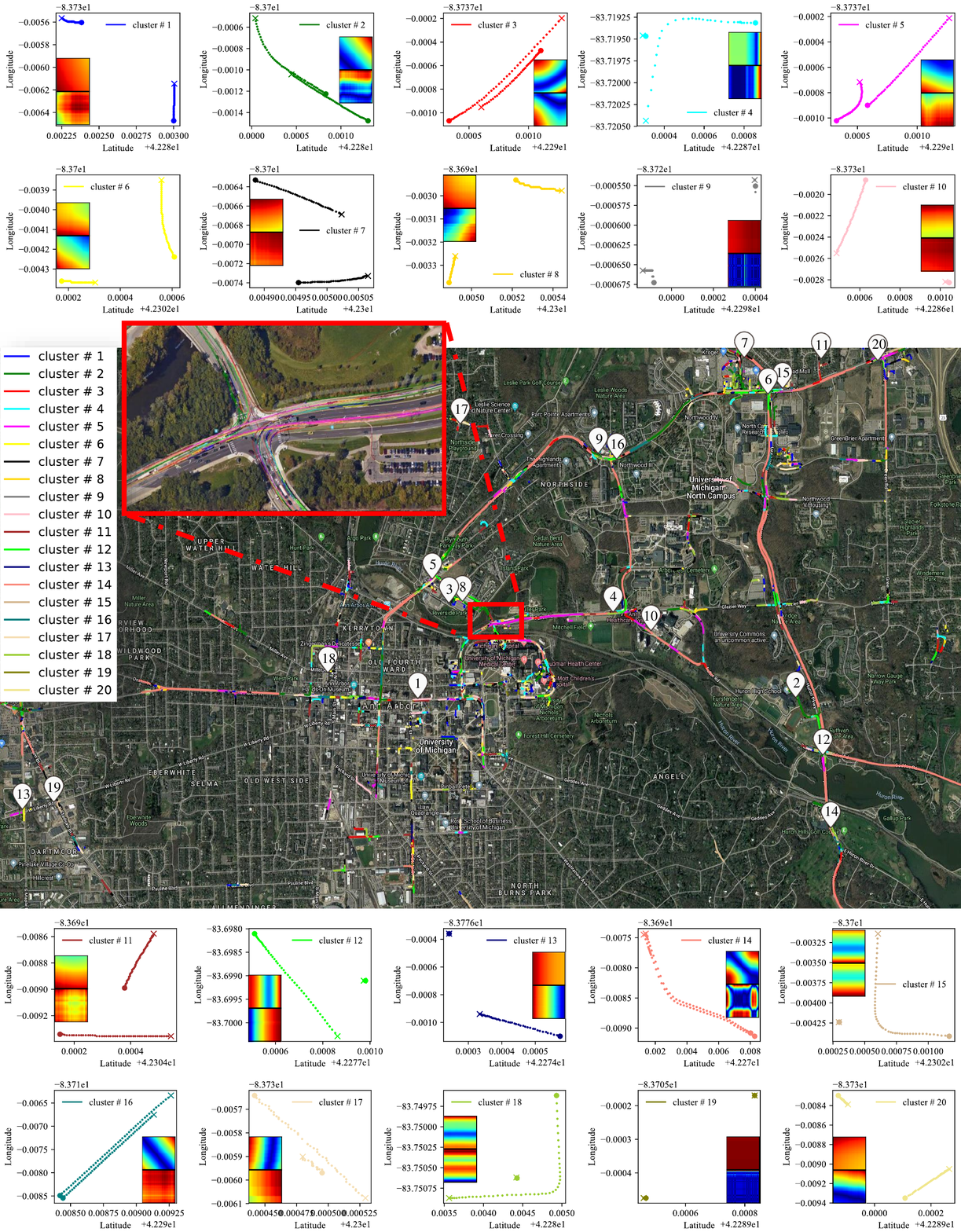}
\caption{GPS data visualized on Google map with 20 typical driving primitives in each cluster. The location markers with number on the map are coordinated with the 20 samples from each cluster.}
\label{fig:gps_all20pri}
\end{figure*}

Fig. \ref{fig:gps_all20pri} visualizes all the driving primitives with the specified colors on the Google map. The driving primitives belong to the same cluster have the identical color. We also randomly select one driving primitive from each cluster (totally 20 clusters) and then draw their trajectories with starting point (dot) and endpoint (cross) as well as also add the heat maps of feature matrices. The top and bottom heat maps in each subplot are corresponding to the trajectories and speeds, respectively. It can be observed that the behavior of driving encounters is most occurring at intersections, which is also one of the most challenging scenarios for intelligent vehicles. Fig. \ref{fig:gps_all20pri} also displays all the 20 typical driving primitives which represent corresponding clusters. According to the extracted driving primitives of encountering behavior, we can make detailed analysis with respect to different types of typical scenarios as follows.

\textit{1) Both two vehicles keep still.} Cluster \#$19$ displays the case wherein two vehicles  keep still over the whole period of time. This is one typical behavior in real traffic such as when two cars stop at intersections with red traffic lights or stop signs. Cluster \#$9$ displays a little different case, which is hard to be recognized as a comprehensible behavior. The trajectories of two vehicles in cluster \#9 are mainly caused by the drifting of GPS sensor. It is hard to tell the in-depth difference between these two clusters. That is partly caused by the characteristics of unsupervised machine learning since the initial state is randomly set.

\textit{2) Two vehicles move in the vertical direction.} Clusters \#$1$, \#$6$, \#$8$, \#$11$ and \#$20$ describe this kind of encountering behavior. In this scenario, clusters \#$1$, \#$6$ and \#$8$ describe the similar behavior that one vehicle is approaching the intersection while the other vehicle is departing from the intersection, but with different speed profiles. More specifically, the speeds of two vehicles remain unchanged in cluster \#$1$ while go up in cluster \#$6$. Cluster \#$8$ displays the behavior that the vehicle is speeding down and approaching the intersection and another vehicle is driving away from the intersection with speed-up. Cluster \#$11$ describes the behavior that one vehicle is driving away from the intersection and the other vehicle is going to pass the intersection, wherein the vehicle passing the intersection keeps a high speed while the vehicle leaving from the intersection keeps a relative low speed. For cluster \#$20$, it describe the case that one vehicle is approaching the intersection while another vehicle is passing the intersection with their relative speed decreasing.

\textit{3) Two vehicles move in the same direction.} Clusters \#$2$, \#$7$, \#$14$ and \#$16$ illustrate this kind of encountering behavior. More specifically, cluster \#$2$ describes the behavior that one vehicle follows the other vehicle while their relative speed is decreasing. Cluster \#$7$ describes the behavior that two vehicles move on the branch road and will meet at a Y-intersection. This cluster is the most common encountering behavior in our case. 462 driving primitives belong to it. The relative distance of two vehicles decreases and two vehicle's speeds remains unchanged. Cluster \#$14$ displays that two vehicles drive in parallel with the same direction, wherein the speed difference of these two vehicles is very small and both of them accelerate first then maintain the speed and decelerate lastly. Cluster \#$16$ shows the similar behavior as cluster \#$14$, except for a little difference in driving at a constant speed.

\textit{4) Two vehicles move in the opposite direction.} This kind of typical behavior can been seen in clusters \#$3$ and \#$17$. In both two clusters, one vehicle's speed is relative low and has related short trajectory. Another vehicle has relative high speed and long trajectory. The relative speed of two vehicles is nearly unchanged during the whole period of time of this driving primitive. The difference lies in the absolute speed. In cluster \#$3$, the speeds of two vehicles go up. And in cluster \#$17$, two vehicles maintain the speed unchanged.

\textit{5) One vehicle moves and the other vehicle is still.} Clusters \#$4$, \#$10$, \#$12$, \#$13$, \#$15$ and \#$18$ represent this behavior. In this case, these clusters can be preliminary separated as two classes: the moving vehicles in clusters \#$4$, \#$15$ and \#$18$ make a turn and the moving vehicles in clusters \#$10$, \#$12$ and \#$13$ make a straight movement. For turning behavior, cluster \#$4$ describes a left turn (in the plot's coordinates), wherein the speed of the moving vehicle increases from zero and accelerates in the whole period of time during this driving primitive. Correspondingly, clusters \#$15$ and \#$18$ describe a case of right turn behavior. The difference between the cluster \#$15$ and the cluster \#$18$ lies in the speed variance situation. In cluster \#$15$, the moving vehicle speeds down to make a turn and then speeds up, while in cluster \#$18$, the moving vehicle speeds down to make a turn and decelerates further after that.

\textit{6) The front vehicle moves straightly and the behind vehicle moves straightly first and then makes a turn.} As shown in cluster \#$5$, this behavior usually happens at intersections, wherein the speed of the behind vehicle is lower than that of the front vehicle and both speeds of two vehicles do not change a lot during the whole period of time of this primitive.

%\textit{7) Others.} Cluster \#$9$ displays the case different from the aforementioned cases, which is hard to be recognized as a comprehensible behavior, but little similar to cluster \#$19$, which describe the behavior that both two vehicles keep still. The trajectories of two vehicles in cluster \#9 are mainly caused by the drifting of GPS sensor. It is hard to tell the in-depth difference between these two clusters. That is partly caused by the characteristics of unsupervised machine learning since the initial state is randomly set. %Thus, it may separate some similar features into different clusters.

%Cluster $2$ displays the circumstance which occurs at the intersection. Two vehicles gradually driving away with each other with perpendicular direction at the intersection. The speed of two vehicles doesn't change a lot during the whole 5.00s time span. Cluster $20$ also occurs at the intersection. In this case, two vehicles encounter each other. One car doesn't move during the whole time span, the other car goes straight and turns right. Both two cases described above are very common in the real driving situation, which is consistent with common sense.

%\textit{2) Two most uncommon driving primitives.} Cluster $11$ and Cluster $13$ display two most uncommon driving primitives. They are similar to each other, which both describe the situation one vehicle moves and the other doesn't move all the time. The difference is that the moving vehicle in cluster $11$ accelerates from zero. On the contrary, th moving vehicle in cluster $13$ decelerates to zero. 

\subsection{Further Discussion}
In this paper, we mainly proposed a unsupervised learning approach to extract driving primitives from driving encounters and then cluster them. The general steps include driving primitive extraction, feature representation and clustering. However, there still exist some challenging problems during these steps, discussed as follows.

\textit{1) The quality of raw data.} All the data we use are from SPMD, which is collected by normal on-board GPS and speed sensors, which is low accuracy and might be unreliable. Considering unexpected factories such as weather, signal interference, etc., the noises and distortion can be unavoidable. Although we get satisfied results by applying aforementioned methods in our case, how to collect high-quality raw data is still a huge challenge, which is mainly caused by the limitation of sensors. With investigating deeply, the adverse effect of this problem may increase. Therefore, how to evaluate and modify raw data to remove aberrant data will be one of our future work. 

\textit{2) The criterion of clustering.} In our case, $k$-means was implemented to cluster driving primitives. However, the difficulty of determining the number of clusters and accuracy of clustering results are the drawbacks of unsupervised machine learning. In our case, we introduce within distance $\lambda_w$ and between distance $\lambda_b$ to decide the number of clusters and tell the quality of clustering results. Based on that, we took a reasonable selection by considering the quality of clustering result and computing cost. However, there still exist some clusters which are hard to interpret for human. We would further focus on the measurement of clustering results over the feature space. Besides, our feature space is 5000 dimensional, which is very huge; therefore, how to select key components which can decrease the dimensions as well as keep useful features is also challenging. In this paper, we keep all 5000 feature components since we intend to keep as many effective features as we can. However, with the increase of data amount, the computing cost will increase exponentially. Thus, improving the computing efficiency of clustering high-dimensional data will be taken into future work.

\section{Conclusion}

In this paper, we presented a driving primitive-based framework to gain an insight into complicated driving encounters. We implemented a nonparametric Bayesian learning (NPBL) approach to extract the basic build blocks (i.e., driving primitives) of driving encounters. The experiment results from naturalistic driving data demonstrate that the behavior of driving encounters can be decomposed into finite kinds of driving primitives. In order to make analysis easily, we introduced a distance-based feature measurement approach to measure the similarity or divergence of driving primitives in both spatial and temporal spaces. Finally, we found that the driving encounters with two vehicles engaged can be fundamentally decomposed into 20 kinds of primitives. It has an inspiring influence on systematically understanding the behavior of driving encounters, enabling us to analyze and design rare but risky scenarios. Thus, it has great potential to be implemented to evaluate the safety of autonomous vehicles. What's more, the information of driving primitives can provide a reasonable basis for modeling decision-making process when two vehicle encountering with each other in complex scenarios. 

\section*{Acknowledgment}
\label{sec: ack}

Toyota Research Institute (``TRI") provided funds to assist the authors with their research but this article solely reflects the opinions and conclusions of its authors and not TRI or any other Toyota entity.

\ifCLASSOPTIONcaptionsoff
  \newpage
\fi



\bibliographystyle{IEEEtran}
\bibliography{ref}


% \end{thebibliography}


\begin{IEEEbiography}[{\includegraphics[width=1in,height=1.25in,clip,keepaspectratio]{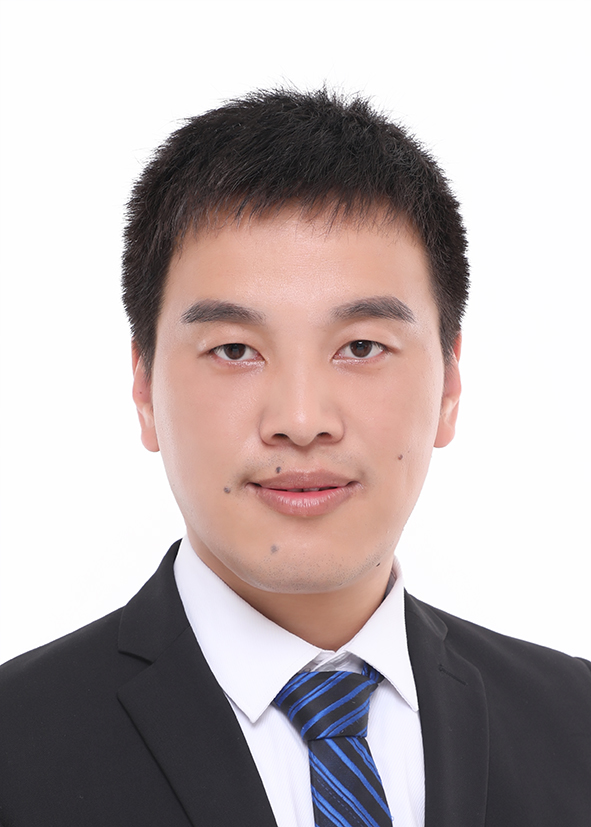}}]{Wenshuo Wang} received his
Ph.D. degree for Mechanical Engineering, Beijing
Institute of Technology (BIT) at June 2018. He is now working as a PostDoc at the Carnegie Mellon University (CMU), Pittsburgh, PA. He also worked as a Research Scholar at the Department of Mechanical Engineering, University of California at Berkeley (UCB) from September 2015 to September 2017 and at the Department of Mechanical Engineering, University of Michigan (UM), Ann Arbor, from September 2017 to July 2018. His research interests include nonparametric Bayesian learning, driver model, human-vehicle
interaction, recognition and application of human driving characteristics.
\end{IEEEbiography}

\begin{IEEEbiography}[{\includegraphics[width=1in,height=1.25in,clip,keepaspectratio]{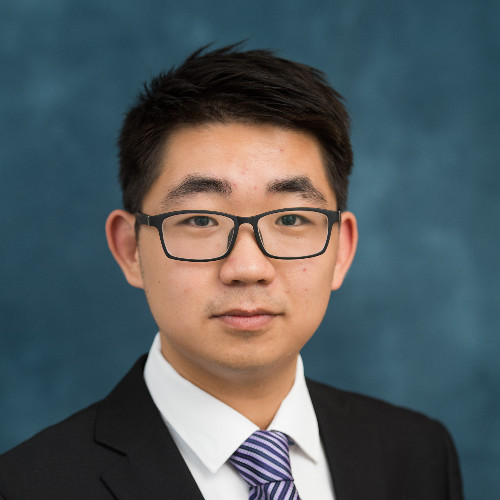}}]{Weiyang Zhang} received his B.S. degree from Zhejiang University at 2017. He now is a Master student with the Department of Mechanical Engineering, University of Michigan, Ann Arbor, MI, USA. His research focuses on autonomous driving, big data analysis, and nonparametric Bayesian learning.
\end{IEEEbiography}

%\newpage

\begin{IEEEbiography}[{\includegraphics[width=1in,height=1.25in,clip,keepaspectratio]{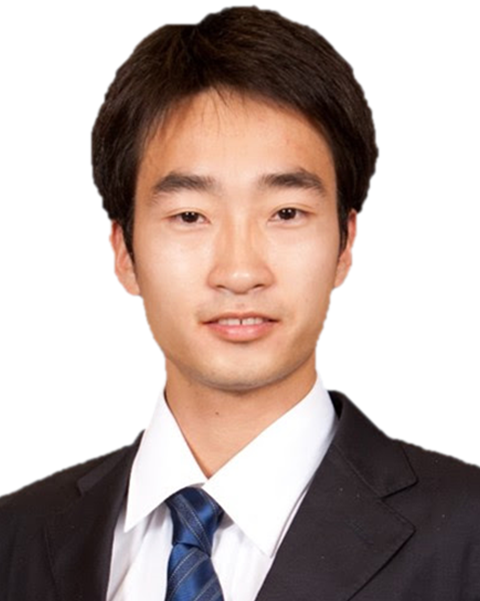}}]{Ding Zhao} received his Ph.D. degree in 2016 from
the University of Michigan, Ann Arbor. He is currently an Assistant Professor at Department of Mechanical Engineering, Carnegie Mellon University. His research focuses on the intersection of robotics, machine learning, and design, with applications on autonomous driving, connected/smart city, energy efficiency, human-machine interaction, cybersecurity, and big data analytics.
\end{IEEEbiography}


%\vfill


\end{document}